# Path planning and Obstacle avoidance approaches for Mobile robot

Hoc Thai Nguyen[1], Hai Xuan Le[2]

[1] Department of Networked Systems and Services, Budapest University of Technology and Economics, Budapest, Hungary

[2] Hanoi University of Science and Technology, Hanoi, Viet Nam

## Abstract

A new path planning method for Mobile Robots (MR) has been developed and implemented. On the one hand, based on the shortest path from the start point to the goal point, this path planner can choose the best moving directions of the MR, which helps to reach the target point as soon as possible. On the other hand, with an intelligent obstacle avoidance, our method can find the target point with the near-shortest path length while avoiding some infinite loop traps of several obstacles in unknown environments. The combination of two approaches helps the MR to reach the target point with a very reliable algorithm. Moreover, by continuous updates of the onboard sensors' information, this approach can generate the MR's trajectory both in static and dynamic environments. A large number of simulations in some similar studies' environments demonstrate the power of the proposed path planning algorithm.

## 1. Introduction

Nowadays, Mobile Robot (MR) is widely used in various fields, such as military [4], industrial [5], agricultural [2, 6], and many other applications [1, 7]. One of the most concern issues in the field of robotics is Robot path planning, which helps the MR to travel from the start point to the target point under obstacle constraints while achieving the shortest path with minimal energy consumption, and the lowest running time. In facts, there are two types of the MR path planning [8]: global path planning and local path planning. The Global path planning approach [9-10] is known as a static motion planning, by which, the trajectory of the MR is computed before the MR starts its motion. The MR's trajectory is generated when the environment is well-known and the terrain is static (no dynamic obstacles). Although this motion planning type ensures that whether the goal point will be reached or it helps to know the target point is unreachable. Unfortunately, two conditions, i.e., known environment, and a static terrain, when applying global path planning type only occur in ideal conditions. It is not easy to know a reliable map of the obstacles, and it is not sure that these obstacles will be static while they are affected by many other environment parameters and events occurrence. The other type, called local path planning [11-13], is a dynamic motion planning approaches, by which the MR's trajectory is generated online based on the current information sensed by onboard sensors of the MR. This motion planning method is known as a more flexible and reliable method than the former type because its quick response to changing of the shapes and the obstacles' positions in the dynamic environment. As we know, autonomous navigation of the MR is necessary for many applications such as rescue robots for searching and rescuing victims inside collapsed building during the disasters [12]. Additionally, the global path planning is not available in some situations when some events suddenly occur on the fixed trajectory of the MR. Thus, the local path planning is an unavoidable choice in these situations. We found, however, the local path planning method still has its limitations: (i) Without global information environment, the global convergence to the goal point is not guaranteed by the local path planner [14]. (ii) Therefore, it may be stuck at some "*local minima points*" and need to recalculate of waypoints during the movement of the MR [15].

To overcome these limitations of the above path planning types, we propose a new approach, namely the Near-shortest path for Mobile Robot (NSPMR) algorithm, which based on the local path planning, to find the nearest shortest-path of the MR, while achieving minimal energy consumption, and lowest runtime of the MR.

Compare with some similar traditional methods, NSPMR algorithm has some advantages and they are highlighted as follows:

- There is no need the global information environment to achieve the near optimal path for a mobile robot.
- NSPMR algorithm can avoid the "*local minima point*" and break the infinite loops [22] along the MR's trajectory to find the goal point if it is a reachable point.

- It is very realistic and useful algorithm when the MR can travel on a narrow road with many roadside obstacles by NSPMR algorithm.

The remainder of the paper is organized as follows:

- The next section summarizes the related works, which try to find the MR's path under obstacles constraint.
- Section 3 provides the system model and assumption of the MR's environment.
- In Section 4, we propose the near optimal path of the MR.
- In Section 5 we give an extensive performance analysis of the proposed algorithms.
- In Section 6, some conclusions are drawn mentioning some future research directions, as well.

## 2. Related works

In the literature, a large number of works on path planning, and obstacle avoidance for the MR have been conducted. In this Section, we introduce some of these researches, which relate to our study.

Bug algorithms [16-17] are known as the simplest obstacle avoidance algorithms. In Bug1 algorithm, in order to pass the obstacles $OB_i$, robot walks along the obstacle boundary from the "*hit point*" ($H_i$ - the first point when robot hit the obstacle $OB_i$) to find the "*leave point*" ($L_i$), which has the shortest distance to the goal point first, then from the hit point $H_i$ robot goes back to the "*leave point*" $L_i$ before leaving to the goal point. This approach is depicted in Figure 1.

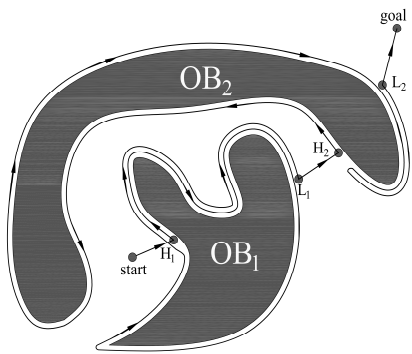

Fig. 1 Mobile robot's path by Bug1 algorithm with $H_1$, $H_2$, hit points, and $L_1$, $L_2$, leave points [17]

Although the Bug1 algorithm guarantees that the robot can reach any target points if they are reachable goal points. However, it is an inefficient algorithm with long boundary obstacles. In these cases, the robot may waste time and energy for travelling around the obstacle boundary.

To overcome this limitation of the Bug1 algorithm, the robot will leave the obstacle $OB_i$ as soon as the "*leave point*" $L_i$ is found in Bug2 algorithm, as shown in Figure 2. It is not necessary for the Robot to move along $H_iL_i$ arc two times as in Bug1 algorithm, therefore the length of the MR's trajectory by Bug2 algorithm will be shorter, of course, saving the running time, and energy consumption for the robot.

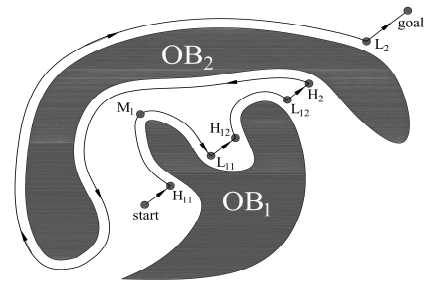

Fig. 2 Mobile robot's path by Bug2 algorithm with $H_{11}$, $H_{12}$, $H_2$, hit points, and $L_{11}$, $L_{12}$, $L_2$, leave points [17]

However, Bug2 algorithm is still a non-optimal algorithm, when the "*leave point*" ($L_{11}$) of the $OB_1$ does not locate at the best position ($M_1$), therefore it still has some unnecessary arcs in the MR's trajectory. Or as stated in [14], Bug algorithms have not utilized all the available sensory data to find the shortest path for the MR. Additionally, Bug2 algorithm may get some bad "*leave point*", which lead the MR enter to some infinitive loops. As shown in Figure 3, the MR may not reach the goal point if the "*leave point*" of the obstacle $OB_2$ is $L_{21}$. Thus, it is very important for one motion planning algorithm to avoid these traps of some obstacles.

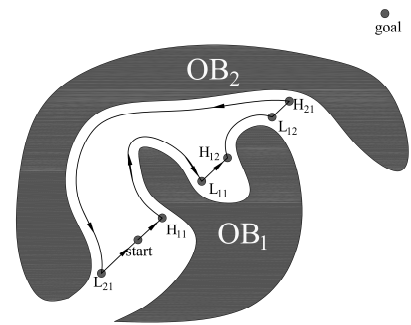

Fig. 3 Infinitive loop in Mobile robot 's trajectory by Bug2 algorithm

Expanding Bug algorithms and most similar to our study, Kamon, I., Rivlin, E. and Rimon, E., 1996 proposed TangentBug algorithm [14], which combines local planning in finding the locally optimal direction of the MR in order to find the shortest path based local Tangent graph (LTG)) with global planning in increasing the maximal direction range of the sensors. The simulation results show that the longer of the sensor range is, the shorter of the robot's average path length is. Comparing with VisBug algorithm [16], the TangentBug algorithm always gives a shorter path length of the robot. However, the leaving conditions in TangentBug algorithm only are held, if the obstacles lie in the visible set of Robot's sensors. It may be failed with some unpredicted and hidden obstacles along robot trajectories [19]. With wrong leaving node ($V$) in the current local tangent graph, the TangentBug algorithm may also hit the infinitive loop as depicted in Figure 3.

Another extension of Bug algorithms is proposed in [18], called PointBug algorithm, which tries to reduce the traveled path length of the robot by detecting some "*sudden points*" and then the robot can follow them to reach the target point. However, for finding these "*sudden points*", the authors assumed that the sensing range is infinitive, which is not available in the real environments. Furthermore, in some real situations, the PointBug algorithm chooses the further "*sudden points*" for the next step of the MR, if it appears earlier on the sensor's scanned direction (left to right or right to left) than some closer sudden points to the target point. Figure 12*d* illustrated this limitation of the PointBug algorithm. As shown in Figure 12*d*, instead of choosing O4 for the next stop from current point O3, it is better for the MR if it chooses O5 (the closer point to the target). To overcome these limitations and improve PointBug algorithm, Farouk Meddah, and Lynda Dib in [23] proposed P-Star algorithm, which finds the robot trajectory in unknown environments by onboard sensors. With some modifications in finding "*sudden points*", P-Star algorithm can reach the target point in some worst cases, which PointBug algorithm fails to find the robot trajectory.

One of the most interesting ideas to solve this problem and break these infinite loops in order to reach the goal point quickly is fuzzy logic controller [19-20, 24-25]. In [19], Xi Li and Byung-Jae Choi utilized a fuzzy logic system with 24 control rules to avoid obstacles and make path planning for a MR in unknown environments. In that study, the authors used ultrasonic sensors to detect the distance from robot to the obstacles during path planning for a mobile robot. However, these fuzzy logic approaches with a large number of control rules may increase the taken time of the MR.

To solve these problems, we propose NSPMR algorithm, which auto detects and avoids the obstacles with local information from robot's sensors and the MR can break any infinite loop along its trajectory in order to achieve the shortest path from start point to the goal point.

# 3. System Model and Assumption

Before presenting our algorithms, we would first like to specify the general assumptions about the model we use in this paper.

## 3.1 System Environments and Assumptions

- The environment is a 2D plane and the target point is a reachable point.
- In the environment (interested area), there are some obstacles, which is a closed curve with finite length and free boundary. The number of obstacles is also finite.
- The MR has no prior knowledge of environment parameters such as locations, shapes, and sizes of the obstacles.
- By equipping GPS and Compass modules [21], the MR can know its current position, moving directions and traveled distance. Before starting its motion, the MR knows the target location, which helps the MR can find the desirable path to the target point $S_k(x_k, y_k)$ from its current position $(x_M(t), y_M(t))$ at the iteration $t$.
- The MR has a large memory enough to store all the points which it passed before and the moving directions which it used at these points.

## 3.2 Network model

Let us assume that the MR needs to move from start point $(x_0, y_0)$ to the target point $S_k(x_k, y_k)$ through a series of vertices $P = \{P_0, P_1, ..., P_k\}$, where $P_0$ denotes start point, and $P_k$ is target point. One feasible and optimal path solution if the path length $\Re$ of the MR is as short as possible.

$$\Re = \min\left\{\sum_{i=0}^{k-1} d(P_i P_{i+1})\right\} \qquad (1)$$

Where $d(P_i P_{i+1})$ is the distance between vertex $P_i$ and its adjacent vertex $P_{i+1}$.

The problem here is the MR may face several obstacles, which lead the MR to hit some local minima points. Additionally, the MR may enter to infinite loops along its trajectory by some shapes of the obstacles. Our path planning is to find the shortest path of the MR while avoiding these obstacles and reach the reachable target point.

# 4. The Proposed Algorithms

## 4.1 The path planner of the mobile robot

After estimating the current position, the MR can communicate with the Base Station to receive the target location, where it needs to move to finish some special tasks (e.g. Fire-fighting [1], spraying task for Agricultural applications [2]) with the smallest energy consumption and under the obstacles constraint of the sensing field. For simplicity, without loss of generality, the current position of the MR is specified by x-axis and y-axis coordinates, and the real angular movement $\theta(t)$ of the MR at iteration

$t$. And it is given by $MR = \begin{bmatrix} x_M(t) \\ y_M(t) \\ \theta(t) \end{bmatrix}$.

We assume that $\theta_d(t)$, is the desired angular movement of the MR at iteration $t$, which is calculated by Equation 2.

$$\theta_d(t) = \arctan\left(\frac{y_k - y_M(t)}{x_k - x_M(t)}\right) \qquad (2)$$

where $S_k(x_k, y_k)$ is the target point. It is easy to prove that the shortest path of the MR is the straight-line segment connecting the current position of the MR and the target point $\theta(t) = \theta_d(t)$. Unfortunately, there are some obstacles which make the MR cannot move along the straight line. Therefore, it is a better trajectory of the MR, if it is close to the ideal straight line and minimizing the angular $\min\{|\theta(t) - \theta_d(t)|\}$. In order to obstacle avoidance, the MR has 8 sensors $\{I_1, ..., I_8\}$ in 8 different directions, which measure the distance between the current position of the MR and the obstacles. The MR is also equipped with the Swedish wheel type as depicted in Figure 4.

With this wheel type, the MR does not turn its body while changing its moving direction. It means that sensor $I_1$ always senses the East area of the MR. The Boolean output of the sensor will be 1 if there is no obstacle within its sensing range and it is a valid direction of the MR if the MR moves by this sensor's direction. And the Boolean output of the sensor will be 0 otherwise.

$$I_i \atop i=1...8 = \begin{cases} 1 \text{ if there is no obstacle} \\ 0 \text{ otherwise} \end{cases} \qquad (3)$$

From current position of the $MR(x_M, y_M)$, if there is no obstacle within the sensor range, the MR can move one of 8 adjacent points, as shown in Figure 5. Where $\delta$ is the MR's length, and it should be shorter than the distance measurement of the onboard sensor $d$, which helps the MR's movement without hitting the obstacle boundary.

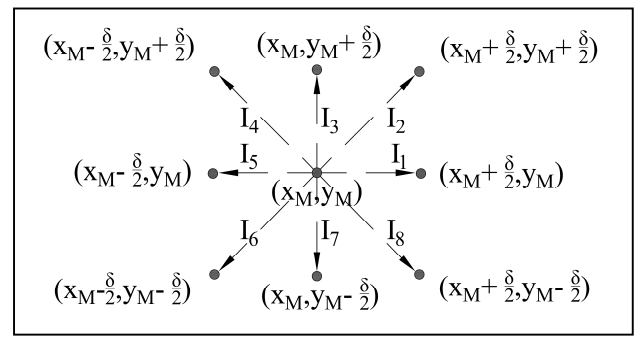

Fig. 5 Adjacent points to the current point of the MR

The moving direction of the MR will be chosen by the following equation:

$$\theta(t) := \min_{\theta_i}\{|\theta_i - \theta_d(t)|\} \qquad (4)$$

where $\theta_i$ is chosen one of 8 cases, given in Table 1.

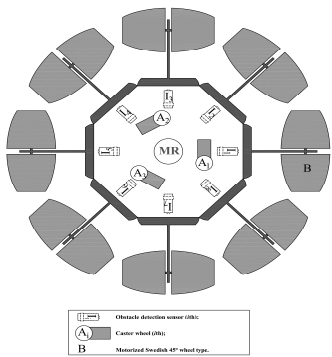

Fig. 4 The locomotion mechanism of the MR

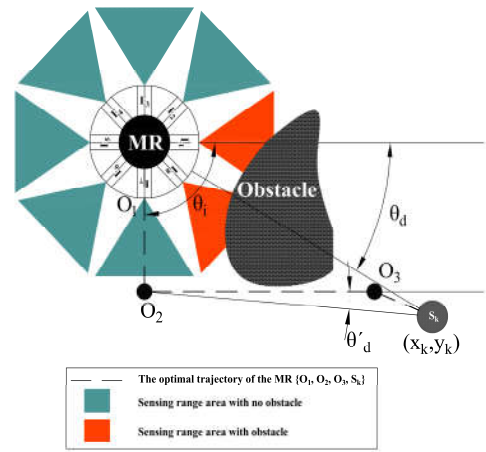

Fig. 6 The optimal trajectory of the MR

Table 1: The choice of the MR's moving direction

| Boolean output of sensors | | | | | | | | | The MR location |
|---|---|---|---|---|---|---|---|---|---|
| $I_1$ | $I_2$ | $I_3$ | $I_4$ | $I_5$ | $I_6$ | $I_7$ | $I_8$ | $\theta_i$ | Next position |
| 1 | - | - | - | - | - | - | - | $0°$ | $(x_M, y_M + \delta/2)$ |
| 0 | 1 | - | - | - | - | - | - | $45°$ | $(x_M + \delta/2, y_M + \delta/2)$ |
| 0 | 0 | 1 | - | - | - | - | - | $90°$ | $(x_M + \delta/2, y_M)$ |
| 0 | 0 | 0 | 1 | - | - | - | - | $135°$ | $(x_M + \delta/2, y_M - \delta/2)$ |
| 0 | 0 | 0 | 0 | 1 | - | - | - | $180°$ | $(x_M, y_M - \delta/2)$ |
| 0 | 0 | 0 | 0 | 0 | 1 | - | - | $225°$ | $(x_M - \delta/2, y_M - \delta/2)$ |
| 0 | 0 | 0 | 0 | 0 | 0 | 1 | - | $270°$ | $(x_M - \delta/2, y_M)$ |
| 0 | 0 | 0 | 0 | 0 | 0 | 0 | 1 | $315°$ | $(x_M - \delta/2, y_M + \delta/2)$ |
| The current position of the MR $(x_M, y_M)$ | | | | | | | | | |
| "-" indicates the output of the onboard sensor $I_i$ may be 0 or 1. | | | | | | | | | |

Figure 6 depicts the results of one example simulation to find the optimal trajectory of the MR under obstacle constraint. Here, at the current position $O_1$ at the iteration $t$, the Boolean statements of the sensors $I_1$ and $I_8$ are 0 because of the obstacle, while the desired angular $\theta_d(t) = 315°$. Thus, the moving direction of the MR is chosen $\theta(t) = \theta_7 = 270°$, and the MR moves to the next stop $\left(x_M(t-1), y_M(t-1) - \dfrac{\delta}{2}\right)$ at point $O_2$. At the new position ($O_2$), the MR continues to compute the new desired angular at the iteration $t+1$ and $\theta_d^{'}(t+1) = 355°$ and then finding new moving direction of the MR: $\theta_i^{'}(t+1) = \theta_1 = 0°$, and move to the next stop: $\left(x_M(t) + \dfrac{\delta}{2}, y_M(t)\right)$. These steps will be repeated until the MR can reach the source node $S_k$. Figure 7 illustrates the results of another example, while the optimal trajectory of the MR is $\{O_1, O_2, O_3, O_4, O_5, O_6, O_7, S_k\}$. By this technique, the MR can easily detect and avoid some local minima points along its traveled path. Figure 8 illustrates one trajectory generated by our algorithm to solve local minima point problem.

Unfortunately, If the distance measurement of the onboard sensors ($d$) is not long enough, the MR may hit local minima points or continue moving in some infinite loops when it hits some shape types of the obstacle as shown in

Figure 9. Figure 9$a$ depicts one example, when the MR hits the "*local minima point*". In this case, only sensor $I_2$ and $I_8$ can detect the obstacle, while the output of sensor $I_1$ is true.

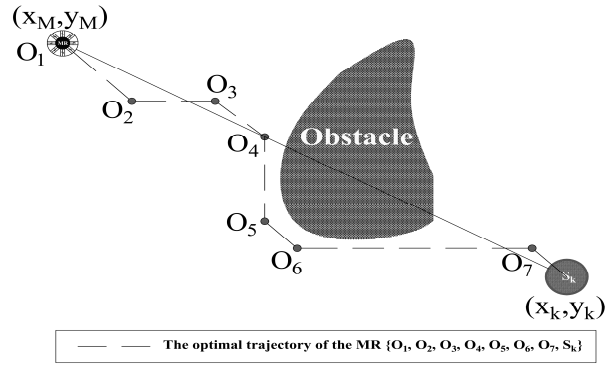

The optimal trajectory of the MR $\{O_1, O_2, O_3, O_4, O_5, O_6, O_7, S_k\}$

Fig. 7 The shortest path from the MR to the source node

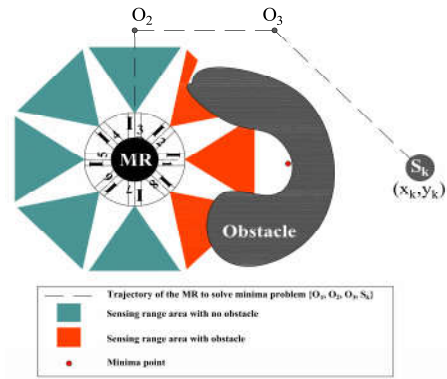

Trajectory of the MR to solve local minima problem $\{O_1, O_2, O_3, S_k\}$
Sensing range area with no obstacle
Sensing range area with obstacle
Minima point

Fig. 8 Trajectory of the MR to solve local minima point problem

CrossMark
click for updates

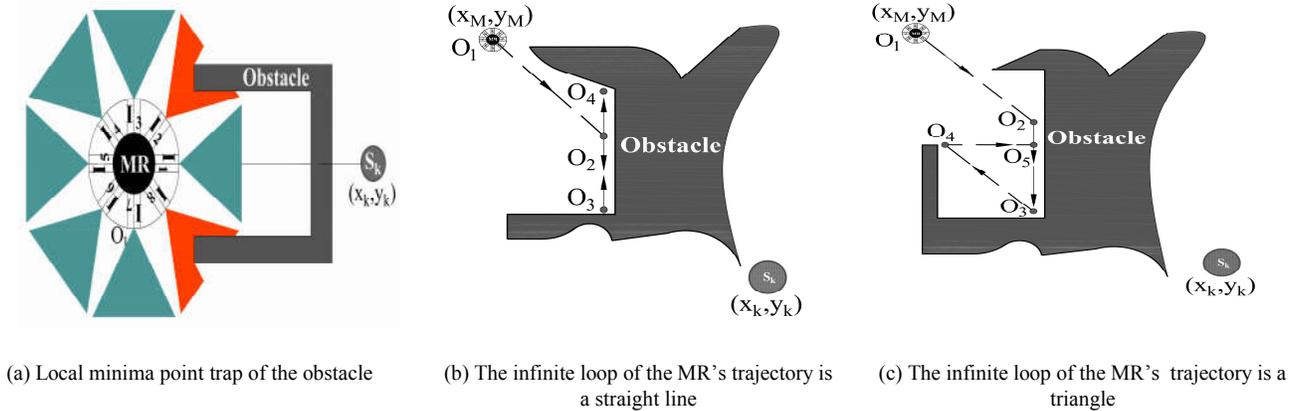

(a) Local minima point trap of the obstacle    (b) The infinite loop of the MR's trajectory is a straight line    (c) The infinite loop of the MR's trajectory is a triangle

Fig. 9 The closed-loop trajectory of the MR under obstacle-constraint

Therefore, the MR continuously keeps its moving direction toward the target point $S_k$ and faces to the local minima problem. In Figure 9$b$, after reaching the $O_2$ the MR moves to the next stop $O_3$ and then $O_4$. The infinite loop $O_3O_4$ may occur if the movement of the MR only depends on 8 sensors. Or in Figure 9$c$, the MR moves from point $O_1$ to point $O_2$ by the ideal straight line from the start point to the target point. At the point $O_2$, under the obstacle constraint, the MR has to move to the point $O_3$. Here, the MR travels along triangle's boundary $O_3O_4O_5$. And it is also the infinite loop trajectory of the MR.

To escape the infinite loop, three following priority moving rules of the MR should be done:

- **Priority moving rule I**: The angular movements of the MR at two adjacent iterations are not opposite directions; As depicted in Figure 9$b$, after moving from point $O_2$ to point $O_3$ at the iteration ($t$), the MR should not move in the opposite direction from point $O_3$ to point $O_2$ or to point $O_4$ at the next iteration ($t+1$).

- **Priority moving rule II**: At each position, the angular movement of the MR will be stored, and it will not be used for the next movement from this position. Thus, by this method, one position with 8 different directions will not be visited by the MR more than 8 times. In the example depicted in Figure 9$c$, $O_5$ will be a break point of the infinite loop. At point $O_5$, after the first loop, the MR will not move to the point $O_3$ or to point $O_4$ again. Instead of this case, the MR will choose 6 other directions if they are valid directions.

- **Priority moving rule III**: In some worse cases, the MR hits local minima points with only one valid direction. In these cases, the Priority moving rule $I$ and $II$ are not available and the MR will only step back its trajectory toward the valid direction, remember these dead points and never moves back again.

To combine these Priority moving rules and the MR's moving direction choice technique, we would like to propose the NSPMR algorithm. The details of the NSPMR are depicted in Figure 10.

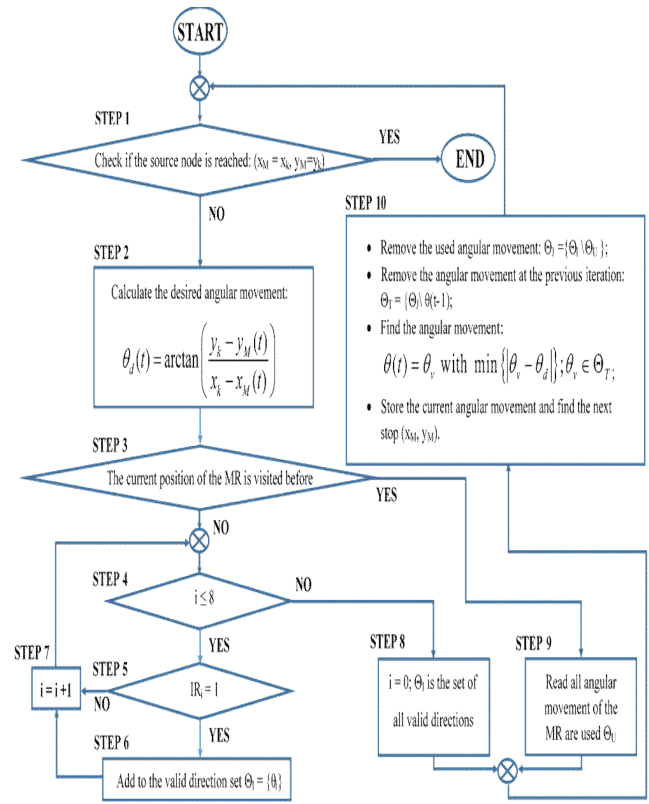

Fig. 10 Flowchart of the NSPMR algorithm

## 5. Performance Study

This Section compares the performance of the proposed NSPMR algorithm with other related studies, Bug1, Bug2 algorithms [17], and approaches [19-20].

### 5.1 The path planner of the mobile robot

In order to compare the power of our proposed NSPMR algorithm with some well-known traditional algorithms, the simulations were carried out by MATLAB software in two different scenarios with the same environments in [18] and [19], respectively. We set some simulation parameters in Table 2.

Table 2: Simulation parameters

| Parameters | Definition | Values |
|---|---|---|
| $N$ | Number of Mobile Robots | 1 |
| $V$ | Speed of the MR | 10 (m/s) |
| $O$ | Start point position | (0,0) |
| $S$ | Target point position | (25,25) |
| $d$ | Sensor range | 1 (m) |

### 5.2 Simulation results

In the first scenario, the workspace is the same environment in [19], which is depicted in Figure 11. In this scenario, the MR needs to move from start point $O(0,0)$ to the target point $S(25,25)$. There are 3 static obstacles, are located in different positions, whose shapes and sizes are given in Table 3.

As can be seen in Figure 11, in all of three algorithms, the MR can avoid the obstacles and reach the target point successfully. The details of the MR's path are given in Table 4. However, the traveled path length and the traveled path time by each algorithm are different.

Table 3: Obstacles' locations in scenario 1

| Obstacle ID | Coordinates of the vertices |
|---|---|
| OB1 | (5.8, 1), (7.5, 1), (7.5, 9.8), (5.8, 9.8) |
| OB2 | (1, 15), (13.5, 15), (13.5, 17.8), (1, 17.8) |
| OB3 | (18.9, 12), (20,2, 12), (20.2, 22), (18.9, 22) |

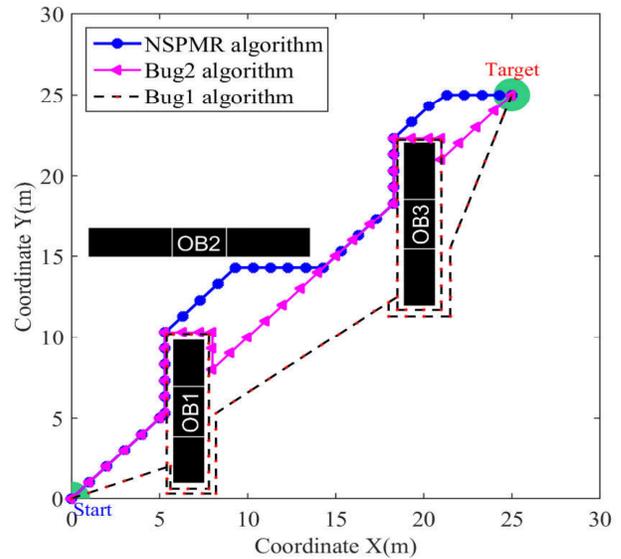

Fig. 11 The MR's path in Experiment [19] to avoid obstacles and reach target point

Table 4: The Coordinates of the vertices on the MR's path

| Methods | The Coordinates of the vertices on the MR's path | Length (m) |
|---|---|---|
| Bug1 | O(0, 0), $A_1$(5.6, 2), $A_2$(5.6, 0.6), $A_3$(7.8, 0.6), $A_4$(7.8, 10.2), $A_5$(5.4, 10.2), $A_6$(5.4, 0.3), $A_7$(8.2, 0.3), $A_8$(8.2, 5.3), $A_9$(18.5, 12.5), $A_{10}$(18.5, 22.2), $A_{11}$(21, 22.2), $A_{12}$(21, 11.7), $A_{13}$(18.5, 11.7), $A_{14}$(18.5, 12.5), $A_{15}$(18, 12.1), $A_{16}$(18, 11.3), $A_{17}$(21.5, 11.3), $A_{18}$(21.5, 15.5), S(25, 25). | 97.1 |
| Bug2 | O(0, 0), $B_1$(5.3, 5.3), $B_2$(5.3, 10.3), $B_3$(8, 10.3), $B_4$(8, 8), $B_5$(18.3, 18.3), $B_6$(18.3, 22.3), $B_7$(21, 22.3), $B_8$(21, 21), S(25, 25). | 45.7 |
| NSPMR | O(0, 0), $C_1$(5.3, 5.3), $C_2$(5.3, 10.3), $C_3$(9.3, 14.3), $C_4$(14.3, 14.3), $C_5$(18.3, 18.3), $C_6$(18.3, 22.3), $C_7$(20.3, 24.3), S(25, 25). | 40.5 |

Table 5 represents the comparison of two factors in 5 algorithms.

Table 5: Comparison of the traveled path length and time of 5 approaches

| Methods | Traveled path length(m) |
|---|---|
| Bug1 | 97.1 |
| Bug2 | 45.71 |
| Approach [19] | 45.6 |
| Approach [20] | 43.8 |
| NSPMR | 40.5 |

It is clear that with more effective techniques, the NSPMR algorithm supports a shorter path length than 4 other algorithms.

Table 6: Comparison of the traveled path length between NSPMR algorithm and Approach [18]

| Methods | Sensor range (m) | Traveled path length (m) |
|---|---|---|
| NSPMR | $d < 2$ | 134.6 |
| | $d = 10$ | 96.8 |
| | $d > 20$ | 78.21 |
| Approach [22] | Unlimited sensor range | 82.5 |

The total length of the trajectory generated by NSPMR algorithm are 58.29%, 11.37%, 11.18%, and 7.53% shorter than Bug1, Bug2, Approach [19], and Approach [20]

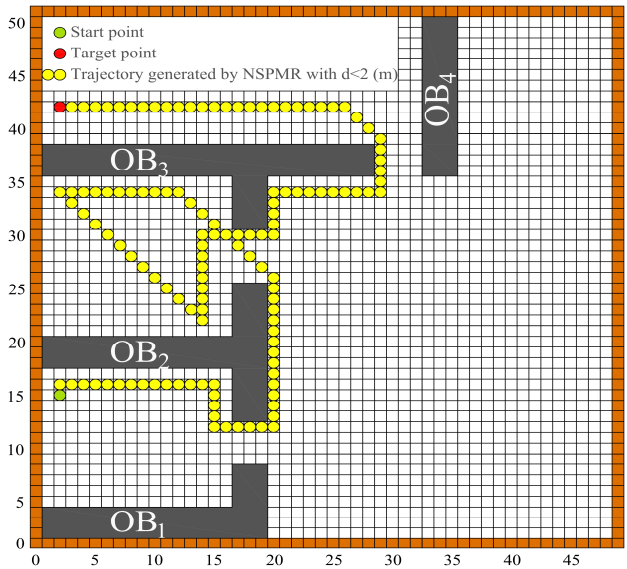

(a) Trajectory generated by NSPMR algorithm with $d < 2$(m)

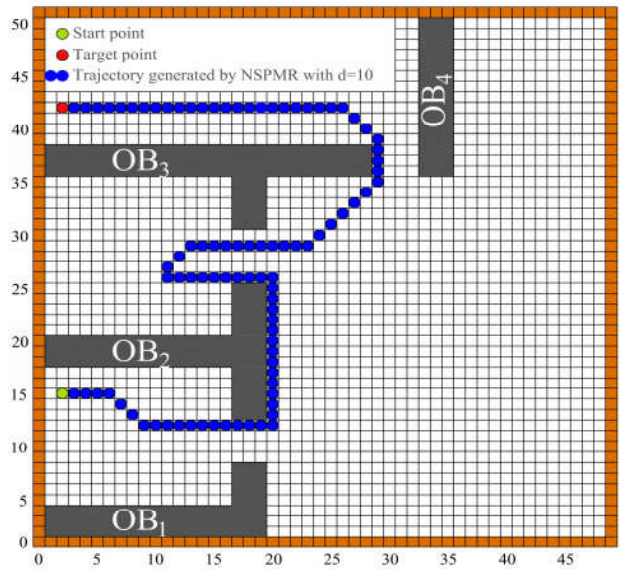

(b) Trajectory generated by NSPMR algorithm with d = 10(m)

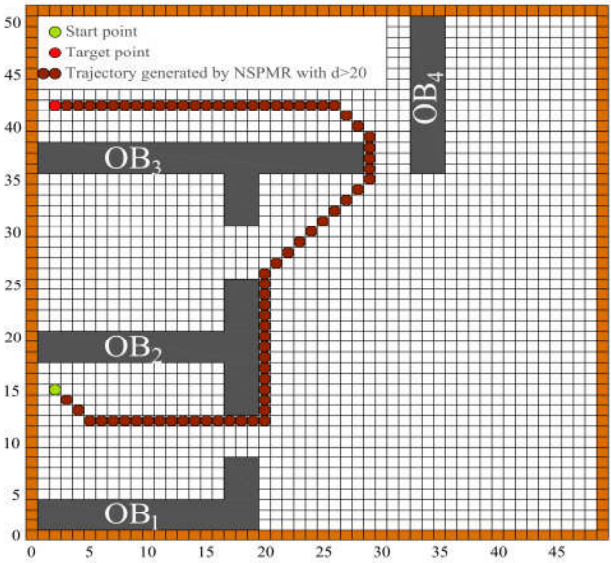

(c) Trajectory generated by NSPMR algorithm with d>20(m)

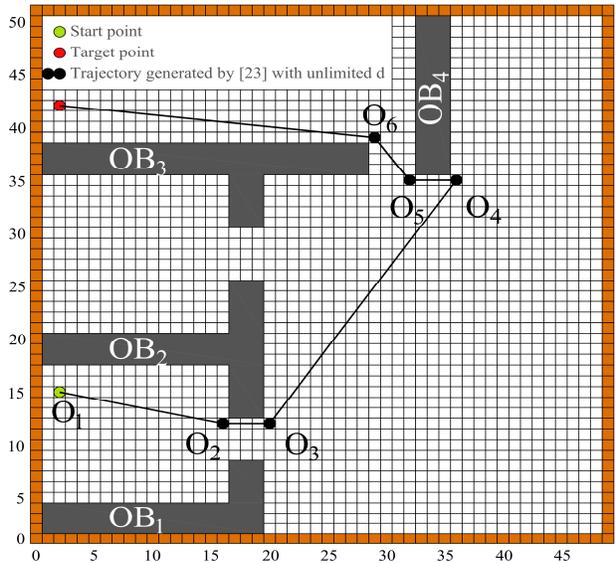

(d) Trajectory generated by PointBug algorithm with unlimited sensor range

Fig. 12 Comparison study between NSPMR and PointBug algorithms

algorithms, respectively. Furthermore, the time taken for the MR reaching target point is also significantly reduced.

In the second scenario, we used "*office like environment*" in [18], by which, we can compare the performance of NSPMR and PointBug algorithms. Figure 12 represents the MR's trajectories generated by these algorithms. As depicted in Figures: 12*a*, 12*b*, and 12*c*, we can see that the longer of the sensing range is, the fewer local minima points the MR hits and the shorter of the MR's traveled paths. When the sensing range is shorter than 2 meters, the MR operated by NSPMR algorithm has a circle segment along its trajectory when it hits obstacle OB$_3$. This circle segment will be disappeared when increasing the sensing range as depicted in Figures 12*b*, and 12*c*. If the sensing range is longer than 20 meters, every points in the rooms can be seen by the MR, and the path planning will be created with the shortest trajectory length. The simulation results are summarized in Table 6. With a flexible path planning, our proposed algorithm can find a shorter traveled path than PointBug algorithm. By NSPMR algorithm, the dynamic path planning has also been implemented in the environments with some dynamic obstacles.

## 6. Conclusions and future directions

In this paper, we present a new approach to making path planning for the MR, which can solve local minima point and infinitive loop problems in existing methods.

The Power of our proposed NSPMR algorithm is demonstrated by simulation results and the comparison studies between our algorithm with some well-known traditional algorithms. It is guaranteed that the MR operated by NSPMR algorithm can reach the target point if it is reachable with a shorter path length and more reliable than some other algorithms. The notable feature of our algorithm is that the MR can avoid the infinite loop problem of some shapes and types of obstacles.

As for future works, we would like to present our experiments in the real world with a greater number of obstacles, and some more dynamic obstacles beside static obstacles.

**Hoc T. Nguyen** received the B.S degree in Electrical and Electronic Engineering from Hanoi Agriculture University in 2006, and the M.A. degree in Automation from Vietnam National University of Agriculture in 2011. He is currently pursuing the Ph.D degree in Infocommunication Technologies at Budapest University of Technology and Economics – Hungary. His research interests include Wireless Sensor Networks, Machine learning, Artificial Neural Network.

**Hai X. Le** received M.A. degree in Electrical and Electronic from Vietnam National University of Agriculture in 2011. He is actually working on Ph.D at Hanoi university of science and technology. His main researches are Non-linear adaptive control, Fuzzy systems – Neural network, Wireless Sensor Networks, Microcontroller Applications and Programmable Logic Controller.